\documentclass[11pt,a4paper]{article}
\usepackage[hyperref]{emnlp2020}
\usepackage{times}
\usepackage{latexsym}
\usepackage{amsmath}
\usepackage{amssymb}
\usepackage{multirow}
\usepackage{enumitem}
\usepackage{colortbl}
\usepackage{pifont}
\usepackage{amsfonts}
\usepackage{floatrow}
\usepackage{graphicx}
\usepackage{subfigure}
\usepackage{xcolor}
\usepackage{color}
\usepackage{lipsum}
\usepackage{caption}
\usepackage{booktabs}
\usepackage{cleveref}
\crefname{section}{§}{§§}
\Crefname{section}{§}{§§}

\definecolor{mygray}{gray}{.9}

\DeclareMathOperator*{\argmax}{argmax}

\usepackage{microtype}

\aclfinalcopy 
\def\aclpaperid{1225} 

\title{Uncertainty-Aware Label Refinement for Sequence Labeling}

\author{Tao Gui\thanks{\ \ Both authors contributed equally.}$\ ^{12}$, Jiacheng Ye$^{*12}$, Qi Zhang$^{12}$, \\ \textbf{Zhengyan Li$^{12}$, Zichu Fei$^{12}$, Yeyun Gong$^3$ and Xuanjing Huang$^{12}$}\\
  $^1$Shanghai Key Laboratory of Intelligent Information Processing, Fudan University \\
  $^2$School of Computer Science, Fudan University\\
  $^3$Microsoft Research Asia \\
  {\tt \{tgui16, yejc19, qz, lizy19, zcfei19, xjhuang\}@fudan.edu.cn} \\ 
  {\tt yegong@microsoft.com}}
  
\date{}

\begin{document}
\maketitle
\begin{abstract}


Conditional random fields (CRF) for label decoding has become ubiquitous in sequence labeling tasks. However, the local label dependencies and inefficient Viterbi decoding have always been a problem to be solved. In this work, we introduce a novel two-stage label decoding framework to model long-term label dependencies, while being much more computationally efficient. A base model first predicts draft labels, and then a novel two-stream self-attention model makes refinements on these draft predictions based on long-range label dependencies, which can achieve parallel decoding for a faster prediction. In addition, in order to mitigate the side effects of incorrect draft labels, Bayesian neural networks are used to indicate the labels with a high probability of being wrong, which can greatly assist in preventing error propagation. The experimental results on three sequence labeling benchmarks demonstrated that the proposed method not only outperformed the CRF-based methods but also greatly accelerated the inference process.
\end{abstract}

\section{Introduction}
Linguistic sequence labeling is one of the fundamental tasks in natural language processing. It has the goal of predicting a linguistic label for each word, including part-of-speech (POS) tagging, text chunking, and named entity recognition (NER). Benefiting from representation learning, neural network-based approaches can achieve state-of-the-art performance without massive handcrafted feature engineering \cite{ma2016end,lample2016neural,strubell2017fast,peters2018deep,devlin2019bert}.

\begin{figure}[t]
\centering
  \includegraphics[width=2.8in]{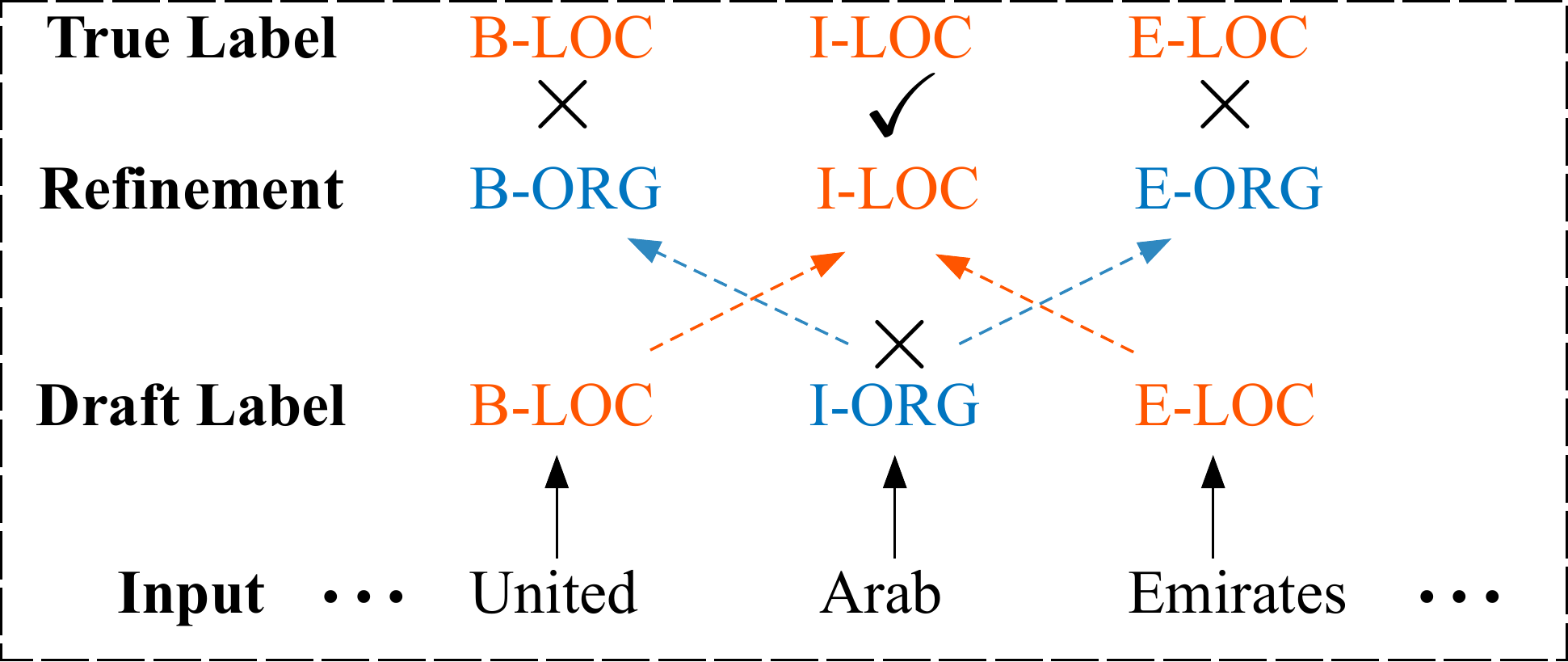}
  \caption{Schematic of label refinement framework \cite{cui2019hierarchically}. The goal is refining the label of ``Arab'' using contextual labels and words, while the refinement of other correct labels may be negatively impacted by incorrect draft labels.} \label{fig:example}
\end{figure}

Although the use of representation learning to obtain better text representation is very successful, creating better models for label dependencies has always been the focus of sequence labeling tasks \cite{collobert2011natural,ye2018hybrid,zhang2018learning}. Among them, the CRF layer integrated with neural encoders to capture label transition patterns \cite{zhou2015end,ma2016end} has become ubiquitous in sequence labeling tasks. However, CRF only captures the neighboring label dependencies and must rely on inefficient Viterbi decoding. Many of the recent methods try to introduce label embeddings to manage longer ranges of dependencies, such as two-stage label refinement \cite{krishnan2006effective,cui2019hierarchically} and seq2seq \cite{vaswani2016supertagging,zhang2018learning} frameworks. In particular, \citet{cui2019hierarchically} introduced a hierarchically-refined representation of marginal label distributions, which predicts a sequence of draft labels in advance and then uses the word-label interactions to refine them.


Although these methods can model longer label dependencies, they are vulnerable to error propagation: if a label is mistakenly predicted during inference, the error will be propagated and the other labels conditioned on this one will be impacted \cite{bengio2015scheduled}. As shown in Figure \ref{fig:example}, the label attention network (LAN) \cite{cui2019hierarchically} would negatively impact the correct predictions in the refinement stage. There are 39 correct tokens that have been incorrectly modified (Table \ref{tab:example}). Hence, the model should selectively correct the labels with high probabilities of being incorrect, not all of them. Fortunately, we find that uncertainty values estimated by Bayesian neural networks \cite{kendall2017uncertainties} can effectively indicate the labels that have a high probability of being incorrect. As shown in Table \ref{tab:example}\footnote{We slightly modified the code using Bayesian neural networks.}, the average uncertainty value of incorrect prediction is 29 times larger than that of correct predictions for the draft labels. Hence, we can easily set an uncertainty threshold to only refine the potentially incorrect labels and prevent side effects on the correct labels.

\begin{table}[t]
\centering
\scalebox{0.9}{
\begin{tabular}{c|c||c|c}
\hline
  \textbf{Draft} & \textbf{Uncertainty} & \textbf{Refinement} & \textbf{\#Tokens} \\
  \hline
  \ding{52} & 0.018 & \ding{52} \ding{217} \ding{56} & \textbf{39} \\
  \ding{56} & \textbf{0.524} & \ding{56} \ding{217} \ding{52} & 54 \\
  \hline
\end{tabular}}
\caption{Results of LAN with uncertainty estimation evaluated on CoNLL2003 test dataset. \ding{52} refers to the correct prediction, and \ding{56} refers to the wrong prediction. We use Bayesian neural networks \cite{kendall2017uncertainties} to estimate the uncertainty. We can see that the uncertainty value of incorrect prediction is 29 times larger than that of correct predictions, which can effectively indicate the incorrect predictions.}
  \label{tab:example}
\end{table}

In this work, we propose a novel two-stage Uncertainty-Aware label refinement Network (UANet). At the first stage, the Bayesian neural networks take a sentence as input and yield all of the draft labels together with corresponding uncertainties. At the second stage, a two-stream self-attention model performs attention over label embeddings to explicitly model the label dependencies, as well as context vectors to model the context representations. All of these features are fused to refine the potentially incorrect draft labels. The above label refinement operations can be processed in parallel, which can avoid the use of Viterbi decoding of the CRF for a faster prediction. Experimental results on three sequence labeling benchmarks demonstrated that the proposed method not only outperformed the CRF-based methods but also significantly accelerated the inference process.

The main contributions of this paper can be summarized as follows: 1) we propose the use of Bayesian neural networks to estimate the uncertainty of predictions and indicate the potentially incorrect labels that should be refined; 2) we propose a novel two-stream self-attention refining framework to better model different ranges of label dependencies and word-label interactions; 3) the proposed parallel decoding process can greatly speed up the inference process; and 4) the experimental results across three sequence labeling datasets indicate that the proposed method outperforms the other label decoding methods.




\section{Related Work and Background}

\subsection{Sequence Labeling}
Traditional sequence labeling models use statistical approaches such as Hidden Markov Models (HMM) and Conditional Random Fields (CRF) \cite{passos2014lexicon,cuong2014conditional,luo2015joint} with handcrafted features and task-specific resources. With advances in deep learning, neural models could achieve competitive performances without massive handcrafted feature engineering \cite{chiu2016named,santos2014learning}. In recent years, modeling label dependencies has been the other focus of sequence labeling tasks, such as using a CRF layer integrated with neural encoders to capture label transition patterns \cite{zhou2015end,ma2016end}, and introducing label embeddings to manage longer ranges of dependencies \cite{vaswani2016supertagging,zhang2018learning,cui2019hierarchically}. Our work is an extension of label embedding methods, which applies label dependencies and word-label interactions to only refine the labels with high probabilities of being incorrect. The probability of making a mistake is estimated using Bayesian neural networks, which will be described in the next subsection.

\subsection{Bayesian Neural Networks}

The predictive probabilities obtained by the softmax output are often erroneously interpreted as model confidence. However, a model can be uncertain in its predictions even with a high softmax output \cite{gal2016dropout}. \citet{gal2016dropout} gives results showing that simply using predictive probabilities to estimate the uncertainty results in extrapolations with unjustified high confidence for points far from the training data. They verified that modeling a distribution over the parameters through Bayesian NNs can effectively reflect the uncertainty, and \textbf{Bernoulli Dropout} is exactly one example of a regularization technique corresponding to an approximate variational distribution. Some typical examples of using Bernoulli distribution to estimate uncertainty are Bayesian CNN \cite{gal2015bayesian} and variational RNN \cite{gal2016theoretically}.

Given the dataset $\mathcal{D}$ with training inputs $\mathbf{X}=\{\mathbf{x}_1, \dots, \mathbf{x}_n \}$ and their corresponding outputs $\mathbf{Y}=\{\mathbf{y}_1, \dots, \mathbf{y}_n\}$, Bayesian inference looks for the posterior distribution of the parameters given the dataset $p(\mathbf{W}|\mathcal{D})$. This makes it possible to predict an output for a new input point $\mathbf{x}^*$ by marginalizing over all of the possible parameters, as follows:
\begin{equation}
p(\mathbf{y}^*|\mathbf{x}^*, \mathcal{D}) = \int p(\mathbf{y}^*|\mathbf{W}, \mathbf{x}^*)p(\mathbf{W}|\mathcal{D})\mathsf{d}\mathbf{W}.\label{equ:inference}
\end{equation}

Bayesian inference is intractable for many models because of the complex nonlinear structures and high dimension of the model parameters. Recent advances in variational inference introduced new techniques into the field. Among these, Monte Carlo Dropout \cite{gal2016dropout} requires minimum modification to the original model. It is possible to use the variational inference approach to find an approximation $q^*_\theta(\mathbf{W})$ to the true posterior $p(\mathbf{W}|\mathcal{D})$ parameterized by a different set of weights $\theta$, where the Kullback-Leibler (KL) divergence of the two distributions is minimized. The integral can be approximated as follows:
\begin{equation}
p(\mathbf{y}^*|\mathbf{x}^*, \mathcal{D}) \approx \sum_{j=1}^T p(\mathbf{y}^*|\mathbf{W}_j, \mathbf{x}^*)q^*_\theta(\mathbf{W}_j). \label{equ:post}
\end{equation}
In contrast to non-Bayesian networks, at test time, Dropouts are also activated. As a result, model uncertainty can be approximately evaluated by summarizing the variance of the model outputs from multiple forward passes.

\section{Uncertainty-Aware Label Refinement}
\begin{figure*}[t]
\centering
  \includegraphics[width=5in]{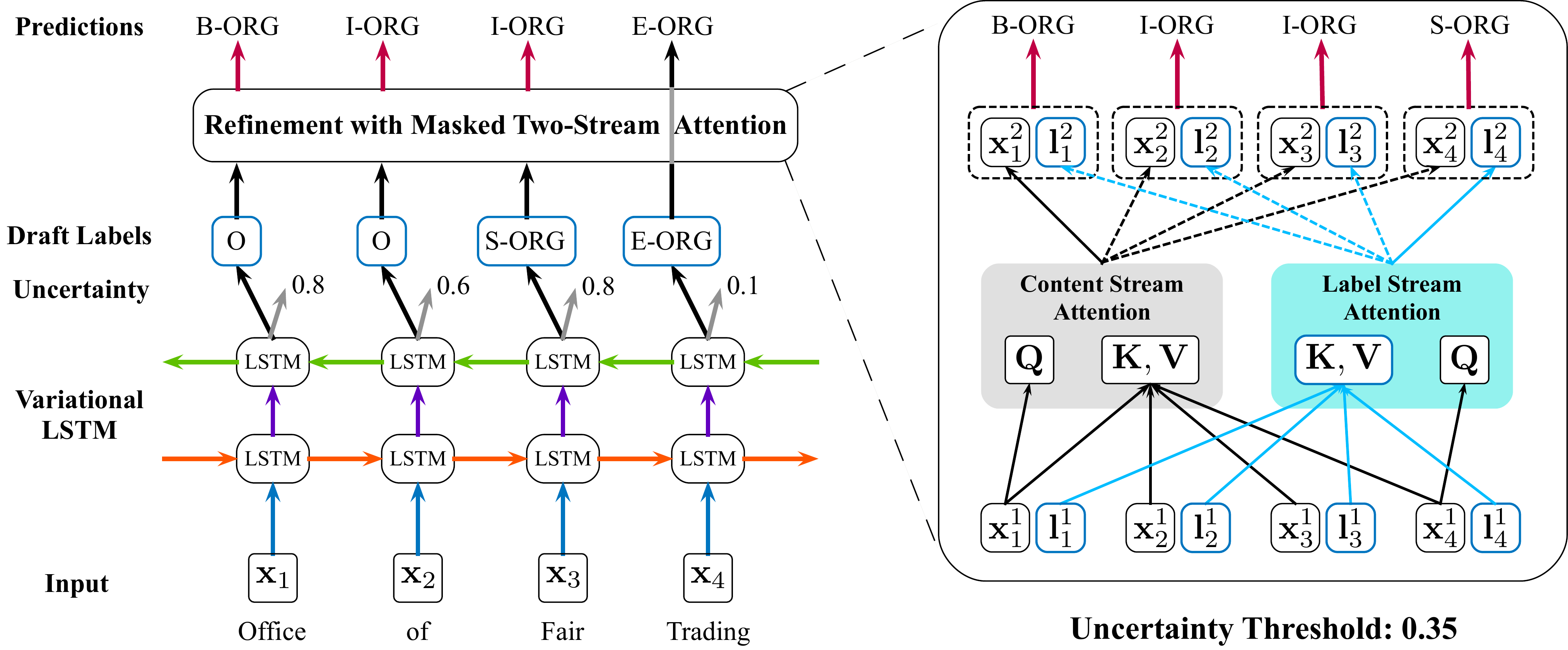}
  \caption{Graphical illustration of architecture and inference process for the proposed UANet. The variational LSTM outputs draft labels and model uncertainties simultaneously. The refinement only works on draft labels with threshold greater than 0.35.} \label{fig:model}
\end{figure*}

In this work, we propose a novel sequence labeling framework, which incorporates Bayesian neural networks to estimate the epistemic uncertainty of the draft labels. The uncertain labels that have a high probability of being wrong can be refined by a two-stream self-attention model using long-term label dependencies and word-label interactions. The proposed model is shown in Figure \ref{fig:model}.

\subsection{Variational LSTM for Uncertainty Estimation}
Long short-term memory (LSTM) stands at the forefront of many recent developments in sequence labeling tasks. To facilitate comparison with LSTM-based models, variational LSTMs \cite{gal2016theoretically} as special Bayesian neural networks are used to encode sentences and determine the labels with a high probability of being wrong. Obviously, the uncertainty estimation methods can also be easily applied to other sequence labeling models, like the CNN and Transformer.

\paragraph{Word Representation} Following \citet{santos2014learning} and \citet{lample2016neural}, we use character information to enhance the word representation. Given a word sequence $S = \{w_1, w_2, \dots, w_n\}$, the product of the one-hot encoded vector with an embedding matrix then gives a word embedding: $\mathbf{w}_i=\mathbf{e}^w(w_i)$, where $\mathbf{e}^w$ denotes a word embedding lookup table. Each word is made up of a sequence of characters $c_1, c_2, \dots, c_l$. We adopt CNNs for character encoding and $\mathbf{x}_i^c$ denotes the output of character-level encoding. Then a word is represented by concatenating its word embedding and its character-level encoding: $\mathbf{x}_i = [\mathbf{w}_i;\mathbf{x}_i^c]$. All the word representations make up an embedding matrix $\mathbf{E} \in \mathbb{R}^{V\times D}$, where $D$ is the embedding dimensionality of $\mathbf{x}$ and $V$ is the number of words in the vocabulary.

\paragraph{Variational LSTM} A common practice of Dropout technique on LSTM is that the technique should be used with the inputs and outputs of the LSTM alone. In contrast, the variational LSTM additionally applies Dropout on recurrent connections by repeating the same mask at each time step. Hence, the variational LSTM can model the uncertainty more accurately.

As shown in Figure \ref{fig:model}, we use the same Dropout vectors $\mathbf{z}_x$ and $\mathbf{z}_h$ on four gates: ``input'', ``forget'', ``output'', and ``input modulation'' as follows:
\begin{equation}
\begin{aligned}
  \begin{bmatrix}
  \mathbf{g}_i\\
  \mathbf{i}_i\\
  \mathbf{f}_i\\
  \mathbf{o}_i\\
  \end{bmatrix}
  &=
  \begin{pmatrix}
    \begin{bmatrix}
    \mathbf{W}^{g}\\
    \mathbf{W}^{i}\\
    \mathbf{W}^{f}\\
    \mathbf{W}^{o}
    \end{bmatrix}
    \bullet
    \begin{bmatrix}
    \mathbf{x}_i' \odot \mathbf{z}_x\\
    \mathbf{h}_{i-1} \odot \mathbf{z}_h
    \end{bmatrix}
  +
    \begin{bmatrix}
    \mathbf{b}^{g}\\
    \mathbf{b}^{i}\\
    \mathbf{b}^{f}\\
    \mathbf{b}^{o}
    \end{bmatrix}
  \end{pmatrix}\\
  \mathbf{c}_i\ \ \ &= \phi(\mathbf{g}_i) \odot \sigma(\mathbf{i}_i) + \mathbf{c}_{i-1}\odot \sigma(\mathbf{f}_i) \\
  \mathbf{h}_i\ \ \ &= \sigma(\mathbf{o}_i)  \odot \phi\left(\mathbf{c}_i  \right),
\end{aligned}
\end{equation}
where $\phi$ denotes the $\mathtt{tanh}$ function, and $\sigma$ is the $\mathtt{sigmoid}$ function. $\odot$ and $\bullet$ represent the Hadamard product and matrix product, respectively. We assume that $t$ is one of $\{g, i, f, o\}$. Then, $\theta = \{\mathbf{E}, \mathbf{W}^t\}$ and the Dropout rate $r$ are the parameters of the variational LSTM.



\paragraph{Draft Labels and Uncertainty Estimation} Assuming that we have completed the training and obtained the optimized approximated posterior $q^*_\theta(\mathbf{W})$ (the optimizing method is shown in \cref{sec:foo}), at inference time, we can predict an output for a new input point by performing Monte Carlo integration in Eq.\ref{equ:post} as follows:

\begin{equation}
\mathbf{p}_i(y=c|S, \mathcal{D}) \approx \frac{1}{M}\sum_{j=1}^M\mathtt{Softmax}(\mathbf{h}_i| \mathbf{W}_j)
\end{equation}
with $M$ sampled masked model weights $\mathbf{W}_j\sim q^*_\theta(\mathbf{W})$, where $q^*_\theta(\mathbf{W})$ is the Dropout distribution. In order to make the model with multiple sampling the same speed as the standard LSTM, we repeat the same input $M$ times to form a batch and run in parallel on the GPU. Hence, $M$ samples can be done concurrently in the forward passes, resulting in constant running time identical to that of standard Dropout \cite{gal2016dropout}, which is verified in Table \ref{tab:speed}.


Similar to classic sequence labeling models, the model applies $y_i^* = \argmax(\mathbf{p}_i)$ to obtain the draft label. Then the uncertainty of this probability vector $\mathbf{p}_i$ can be summarized using the entropy of the probability vector: 
\begin{equation}
u_i = H(\mathbf{p}_i) = -\sum_{c=1}^Cp_c \log p_c.
\end{equation}
In this way, we can obtain the draft labels $Y^*= \{y_1^*, y_2^*, \dots, y_n^*\}$ coupled with the corresponding epistemic uncertainties $U = \{u_1, u_2, \dots, u_n\}$ for each input sentence. We find when the epistemic uncertainty $u_i$ is larger than some threshold value $\Gamma$, then the draft label $y_i^*$ has a high probability of being wrong. Hence, we utilize a novel two-stream self-attention model to refine those uncertain labels using long-term label dependencies and word-label interactions.

\subsection{Two-Stream Self-Attention for Label Refinement}
Given the draft labels and corresponding epistemic uncertainties, we seek the help of label dependencies and word-label interactions to refine the uncertain labels. In order to refine the draft labels in parallel, we use the Transformer \cite{vaswani2017attention} incorporating relative position encoding \cite{dai-etal-2019-transformer} to model the words and draft labels.

In the standard Transformer, the attention score incorporating absolute position encoding between query $q_i$ and key vector $k_j$ can be decomposed as
\begin{equation}
\begin{aligned}
\mathbf{A}_{i,j}^{abs} &= \mathbf{E}_{x_i}^\top\mathbf{W}_{q}^\top\mathbf{W}_{k}\mathbf{E}_{x_j} + \mathbf{E}_{x_i}^\top\mathbf{W}_{q}^\top\mathbf{W}_{k}\mathbf{U}_{j} \\
& + \mathbf{U}_i^\top\mathbf{W}_{q}^\top\mathbf{W}_{k}\mathbf{E}_{x_j} + \mathbf{U}_i^\top\mathbf{W}_{q}^\top\mathbf{W}_{k}\mathbf{U}_{j},\label{absposition}
\end{aligned}
\end{equation}
where $\mathbf{U}\in \mathbb{R}^{L_{max}\times d}$ provides a set of positional encodings. The $i$th row $\mathbf{U}_i$ corresponds to the $i$th absolute position and $L_{max}$ prescribes the maximum possible length to be modeled.

The relative position between labels is very important for modeling the label dependencies. Inspired by \citet{dai-etal-2019-transformer}, we modify the Eq.\ref{absposition} using the relative position encoding to model words and corresponding labels simultaneously, but offer a different derivation, arriving at a new form of two-stream relative positional encodings. We not only provide a word-to-word interactions but also provide a word-to-label interactions correspondence to its counterpart. The relative position encodings are reparameterized as follows:
\begin{equation}
\begin{aligned}
\mathbf{A}_{i,j}^{x2x} &= \mathbf{E}_{x_i}^\top\mathbf{W}_{qx}^\top\mathbf{W}_{kx}\mathbf{E}_{x_j} + \mathbf{E}_{x_i}^\top\mathbf{W}_{qx}^\top\mathbf{W}_{kR}\mathbf{R}_{i-j} \\
& + \mathbf{u}_x^\top\mathbf{W}_{kx}\mathbf{E}_{x_j} + \mathbf{v}_x^\top\mathbf{W}_{kR}\mathbf{R}_{i-j} \\
\mathbf{A}_{i,m}^{x2l} &= \mathbf{E}_{x_i}^\top\mathbf{W}_{ql}^\top\mathbf{W}_{kl}\mathbf{E}_{y^*_m} + \mathbf{E}_{x_i}^\top\mathbf{W}_{ql}^\top\mathbf{W}_{kR}\mathbf{R}_{i-m} \\
& + \mathbf{u}_l^\top\mathbf{W}_{kl}\mathbf{E}_{y^*_m} + \mathbf{v}_l^\top\mathbf{W}_{kR}\mathbf{R}_{i-m}, \ \
\end{aligned}
\end{equation}
where $\mathbf{A}_{i,j}^{x2x}$ and $\mathbf{A}_{i,m}^{x2l}$ denotes the attention from the $i$th word ($x_i$) to the $j$th word ($x_j$) and the $i$th word ($x_i$) to the $m$th label ($y^*_m$), respectively. $\mathbf{R}_{i-j}$ is the encoding of relative distance between position $i$ and $j$, and $\mathbf{R}$ is the sinusoid matrix like \citet{dai-etal-2019-transformer}. $\varphi = \{\mathbf{W}$, $\mathbf{u}$, and $\mathbf{v}\}$ are learnable parameters. 


Equipping the transformer with our proposed relative positional encoding, we finally arrive at the two-stream self-attention architecture. We summarize the computational procedure for one layer with a single attention head here:
\begin{equation}
\begin{aligned}
\mathbf{V}_x &= \mathbf{E}_x \mathbf{W}_x, \mathbf{a}_x = \mathtt{Softmax}(\mathbf{A}^{x2x})\mathbf{V}_x \\
\mathbf{V}_l &= \mathbf{E}_{y^*} \mathbf{W}_l, \mathbf{a}_l = \mathtt{Softmax}(\mathbf{A}^{x2l})\mathbf{V}_l \\
\mathbf{o}_x &= \mathtt{LayerNorm}(\mathtt{Linear}(\mathbf{a}_x) + \mathbf{E}_x) \\
\mathbf{o}_l &= \mathtt{LayerNorm}(\mathtt{Linear}(\mathbf{a}_l) + \mathbf{E}_{y^*}) \\
& \ \ \ \ \mathbf{H}_x = \mathtt{FeedForward}(\mathbf{o}_x) \\
& \ \ \ \ \mathbf{H}_l = \mathtt{FeedForward}(\mathbf{o}_l).
\end{aligned}
\end{equation}

\subsection{Training and Decoding}\label{sec:foo}
There are two networks to be optimized: one is variational LSTM for draft labels and uncertainty estimation, the other is two-stream self-attention model for label refinement. Our ultimate training goal is to minimize the total loss function on the two models: $\mathcal{L}_{total} = \mathcal{L}_1(\theta, r) + \mathcal{L}_2(\varphi)$.

The variational LSTM performs approximate variational inference. We use a simple Bernoulli distribution (Dropout) $q^*_\theta(\mathbf{W})$ in a tractable family to minimize the KL divergence to the true model posterior $p(\mathbf{W}|\mathcal{D})$. The minimization objective is given by \cite{jordan1999introduction}:
\begin{equation}
\mathcal{L}_1(\theta, r) = - \frac{1}{N}\sum_{i=1}^N\log p(y_i|\mathbf{W}_j) + \frac{1-r}{2N}\parallel\theta\parallel^2,
\end{equation}
where $N$ is the number of data points, and $r$ is the Dropout probability to sample $\mathbf{W}_j\sim q^*_\theta(\mathbf{W})$.

For the two-stream self-attention model, we use the concatenation of $\mathbf{H}_x$ and $\mathbf{H}_l$ for the final prediction $\hat{y}_i = f(\mathbf{H}_x, \mathbf{H}_l| \mathbf{E}_x, \mathbf{E}_{y_m^*})$. In particular, we can optimize the model using cross entropy loss as:
\begin{equation}
\mathcal{L}_2(\varphi) = - \sum_{i=1}^N y_i\log \hat{y}_i,
\end{equation}
where $y_i$ is the one-hot vector of the label corresponding to $w_i$. When training is complete, we can obtain the draft labels $Y^* = \{y_1^*, y_2^*, \dots, y_n^*\}$ and corresponding uncertainties $U = \{u_1, u_2, \dots, u_n\}$ from variational LSTM, and refined labels $\hat{Y} = \{\hat{y}_1, \hat{y}_2, \dots, \hat{y}_n\}$ from two-stream self-attention model. To avoid the correct labels being incorrectly modified, we set an uncertainty threshold $\Gamma$ to distinguish which labels should be used, i.e., we use refined labels when $u_i>\Gamma$ and vice versa (as an example, given $u_1>\Gamma$, $u_2 \leq \Gamma$, and $u_n>\Gamma$, decoding labels will become $\{\hat{y}_1, y^*_2, \dots, \hat{y}_n\}$).


\section{Experimental Setup}
In this section, we describe the datasets across different sequence labeling tasks, including two English NER datasets and one POS tagging dataset. We also detail the baseline models for comparison. Finally, we clarify the hyperparameters configuration of our uncertainty-aware refinement network.

\subsection{Datasets}
We conduct experiments on three sequence labeling datasets. The statistics are listed in Table \ref{tab:datasets}.

\noindent\textbf{CoNLL2003.} The shared task of CoNLL2003 dataset \cite{tjong-kim-sang-de-meulder-2003-introduction} for named entity recognition is collected from Reuters Corpus. The dataset divide name entities into four different types: persons, locations, organizations, and miscellaneous entities. We use the BIOES tag scheme instead of standard BIO2, which is the same as \citet{ma2016end}.

\noindent\textbf{OntoNotes 5.0.} English NER dataset OntoNotes 5.0 \cite{weischedel2013ontonotes} is a large corpus consists of various genres, such as newswire, broadcast, and telephone speech. Named entities are labeled in eleven types and values are specifically divided into seven types, like DATE, TIME, ORDINAL.

\noindent\textbf{WSJ.} Wall Street Journal portion of Penn Treebank \cite{marcus-etal-1993-building}, which contains 45 types of part-of-speech tags. We adopts standard splits following previous works \cite{collins-2002-discriminative, Manning:2011:PT9:1964799.1964816}, selecting section 0-18 for training, section 19-21 for validation and section 22-24 for test.

\begin{table}[t]
\scalebox{0.7}{
\begin{tabular}{l c c c c}
\toprule
\textbf{Dataset} & \textbf{\#Train} & \textbf{\#Dev} & \textbf{\#Test} & \textbf{class} \\
\midrule
CoNLL2003 & 204,567 & 51,578 & 46,666 & 17 \\
OntoNotes & 1,088,503 & 147,724 & 152,728 & 73 \\
WSJ     & 912,344 & 131,768 & 129,654 & 45 \\
\bottomrule
\end{tabular}}
\caption{Statistics of CoNLL2003, OntoNotes and WSJ datasets,
where \# represents the number of tokens in datasets. 
The class number of NER datasets is counted under BIOES tag scheme.}
\label{tab:datasets}
\end{table}

\subsection{Compared Methods}
In this work, we mainly focus on improving decoding efficiency and enhancing label dependencies. Thus, we make comparisons with the classic methods that have different decoding layers, such as Softmax, CRF, and LAN frameworks. We also compare some recent competitive methods, such as Transformer, IntNet \cite{xin-etal-2018-learning}, and BERT \cite{devlin2019bert}.


\noindent\textbf{BiLSTM-Softmax.} This baseline uses bidirectional LSTM to reprensent a sequence. The BiLSTM concatenates the forward hidden state $\overrightarrow{\mathbf{h}}_i$ and backward hidden state $\overleftarrow{\mathbf{h}}_i$ to form an integral representation $\mathbf{h}_i = [\overrightarrow{\mathbf{h}}_i; \overleftarrow{\mathbf{h}}_i]$. Finally, sentence representation $\mathbf{H} = \{\mathbf{h}_i, \cdots, \mathbf{h}_n\}$ is fed to softmax layer for predicting.

\noindent\textbf{BiLSTM-CRF.} A CRF layer is used on top of the hidden vectors $\mathbf{H}$ \cite{ma2016end}. The CRF can model bigram interactions between two successive labels \cite{lample2016neural} instead of making independent labeling decisions for each output. In the decoding time, the Viterbi algorithm is used to find the highest scored label sequence over an input word sequence.

\noindent\textbf{BiLSTM-Seq2seq.} To model longer label dependencies, \citet{zhang2018learning} predicts a sequence of labels as a sequence to sequence problem.


\noindent\textbf{BiLSTM-LAN.} The label attention network (LAN) \cite{cui2019hierarchically} introduces label embedding, and uses consecutive attention layers on the label embeddings to refine the draft labels. It achieves the state-of-the-art results on several sequence labeling tasks.

\noindent\textbf{Rel-Transformer.} This baseline model adopts self-attention mechanism with relative position representations \cite{vaswani2017attention,dai-etal-2019-transformer}. 


\subsection{Hyper-parameter Settings}
Following \cite{ma2016end}, we use the same 100-dimensional GloVe embeddings\footnote{http://nlp.stanford.edu/projects/glove/} as initialization. We use 1-layer variational LSTM with a hidden size of 400 to create draft labels. The vanilla dropout after the embedding layer and the variational dropout is set to 0.5 and 0.25, respectively. We use 2 layers of multi-head transformer for WSJ and CoNLL2003 and 3 for OntoNotes dataset to refine the label. The number of heads is chosen from \{5, 7, 9\}, and the dimension of each head is chosen from \{80, 120, 160\} via grid search. We use SGD as the optimizer for variational LSTM and Adam \cite{kingma2014adam} for transformer. Learning rates are set to 0.015 for SGD on CoNLL2003 and Ontonotes datasets and 0.2 on WSJ dataset. The learning rates for Adam are set to 0.0001 for all datasets. F1 score and accuracy are used for NER and POS tagging, respectively. All experiments are implemented in NCRF++ \cite{yang-zhang-2018-ncrf} and conducted using a GeForce GTX 1080Ti with 11GB memory. More details are shown in our codes\footnote{https://github.com/jiacheng-ye/UANet}.

\begin{table}[t]
\scalebox{0.63}{
\begin{tabular}{l|c|c|c}
\toprule
\textbf{Models} & \textbf{CoNLL2003} & $ \textbf{OntoNotes}$ & \textbf{WSJ}  \\
\hline
\citet{chiu2016named} & 90.91 &86.28 & - \\
\citet{strubell2017fast} & 90.54 & 86.84 & - \\
\citet{liu2018empower} & 91.24 & - & 97.53 \\ 
\citet{Chen2019GRNGR} & 91.44 & 87.67 & - \\
\hline
BiLSTM-CRF \cite{ma2016end} & 91.21 & 86.99 & 97.51 \\
BiLSTM-Softmax \cite{yang-etal-2018-design} & 90.77 & 83.76 & 97.51 \\
BiLSTM-Seq2seq \cite{zhang2018learning} & 91.22 & - & 97.59 \\
Rel-Transformer \cite{dai-etal-2019-transformer} & 90.70 & 87.45 & 97.49 \\
BiLSTM-LAN \cite{cui2019hierarchically} & 90.77$^*$ & 88.16 & 97.58 \\
\hline
\cellcolor{mygray}\textbf{BiLSTM-UANet} ($M=8$) & \cellcolor{mygray}\textbf{91.60} & \cellcolor{mygray}\textbf{88.39} & \cellcolor{mygray}\textbf{97.62} \\
\bottomrule
\end{tabular}}
\centering
    \caption{Main results on three sequence labeling datasets. $^*$ indicates the results by running \citet{cui2019hierarchically}'s released code\protect\footnotemark[5].}
  \label{tab:result}
\end{table}
\footnotetext[4]{https://github.com/Nealcly/BiLSTM-LAN}

\section{Results and Analysis}
In this section, we present the experimental results of the proposed and baseline models. We show that the proposed method not only achieves better performance but also has a significant speed advantage. Since our contribution is mainly focused on the label decoding layer, the proposed model can also be combined with the latest pre-trained model to further improve performance.



\subsection{Main Results}

Table \ref{tab:result} reports model performances on CoNLL2003, OntoNotes, and WSJ dataset, which shows that the proposed method not only can achieve state-of-the-art results on NER task but also is effective on other sequence labeling tasks, like POS tagging. The previous methods leverage rich handcrafted features \cite{huang2015bidirectional,chiu2016named}, CRF decoding \cite{strubell2017fast}, and longer range label dependencies \cite{zhang2018learning,cui2019hierarchically}. Compared with these methods, our UANet model gives better results. Benefitting from the strong capability of modeling long-term label dependencies, the UANet outperforms models with the CRF inference layer by a large margin. Moreover, different from the seq2seq and LAN models that also leverage label dependencies, our UANet model integrates model uncertainty into the refinement stage to avoid side effects on correct draft labels. As a result, it outperforms LAN and seq2seq models on all of the three datasets.

\begin{table}[t]
\scalebox{0.7}{
\begin{tabular}{lccc}
\toprule
\textbf{Models} & \textbf{CoNLL2003} & \textbf{OntoNotes} & \textbf{WSJ} \\
\hline
\textbf{BiLSTM-UANet} & \textbf{91.60} & \textbf{88.39} & \textbf{97.62} \\
\textbf{- Label information} & 91.23 & 87.84 & 97.57 \\
\textbf{- Variational LSTM} \\
\quad Rel-Transformer-Softmax & 90.70 & 87.45 & 97.49 \\
\quad Rel-Transformer-CRF & 91.22 & 87.77 & 97.56 \\
\textbf{- Two-stream self-attention} \\
\quad Variational LSTM-Softmax & 90.83 & 87.11 & 97.46 \\
\quad Variational LSTM-CRF & 91.20 & 87.63 & 97.55 \\
\bottomrule
\end{tabular}}
\caption{Ablation study of UANet.}
  \label{tab:ablation}
\end{table}

\begin{table}[t]
\scalebox{0.73}{
\begin{tabular}{l|c}
\toprule
\textbf{Models} & $\mathbf{F_{1}}$  \\
\hline
IntNet-BiLSTM-Softmax \cite{xin-etal-2018-learning} & 91.43 \\
IntNet-BiLSTM-CRF & 91.64 \\
\cellcolor{mygray}\textbf{IntNet-UANet} & \cellcolor{mygray}\textbf{91.80} \\
\hline
BERT-Softmax \cite{devlin2019bert} & 91.62 \\
BERT-CRF & 91.71 \\
\cellcolor{mygray}\textbf{BERT-UANet} & \cellcolor{mygray}\textbf{92.02} \\
\bottomrule
\end{tabular}}
\centering
    \caption{Results on CoNLL2003 test set. We implement BERT for NER task without document-level information. Original result of BERT in \cite{devlin2019bert} was not achieved with the current version of the library. See a discussion in \cite{stanislawek2019named} and the reported results at \cite{zhang2019modeling}.}
  \label{tab:bert}
\end{table}

\subsection{Ablation Study}
To study the contribution of each component in BiLSTM-UANet, we conducted ablation experiments on the three datasets and display the results in Table \ref{tab:ablation}. The results show that the model’s performance is degraded if the draft label information is removed, indicating that label dependencies are useful in the refinement. We also find that both the variational LSTM and two-stream self-attention play an important role in label refinement. Even though we replace any component with the CRF layer, the performance will be seriously hurt.

We also give our model more complex character representations (IntNet) or use the pretrained model (BERT) to replace the Glove embeddings. We fine-tune the BERT for each task. The results are shown in Table \ref{tab:bert}. We find that the contribution of our model and more complex word representations may be orthogonal, i.e., whether or not the UANet uses the IntNet and BERT, our methods have similar improvements, because of better modeling label dependencies.

\begin{table}[t]
\scalebox{0.7}{
\begin{tabular}{l|c|c|c}
\toprule
& \textbf{CoNLL2003} &  \textbf{OntoNotes} & \textbf{WSJ}  \\
\hline
Average Sentence Length & 13 & 18 & 24 \\
\hline
BiLSTM-Softmax & 3,443 & 2,910 & 3,767  \\
BiLSTM-CRF & 1,433 & 950 & 801  \\
BiLSTM-LAN & 949 & 773 & 943 \\
BiLSTM-Seq2seq & 1,084 & 842 & 751 \\
BiLSTM-UANet ($M=1$) & 1,630 & 1,262 & 1,192\\
BiLSTM-UANet ($M=8$) & 1,474 & 1,129 & 1,044\\
\hline
BERT-CRF & 254 & 231 & 189  \\
BERT-UANet ($M=8$) & 335 & 266 & 214  \\
\bottomrule
\end{tabular}}
\centering
    \caption {Comparison of inference speed. $M$ represents for the number of sampling. We show how many sentences the model can process per second.}
  \label{tab:speed}
\end{table}

\begin{figure}[t]
\centering
  \includegraphics[width=3.1in]{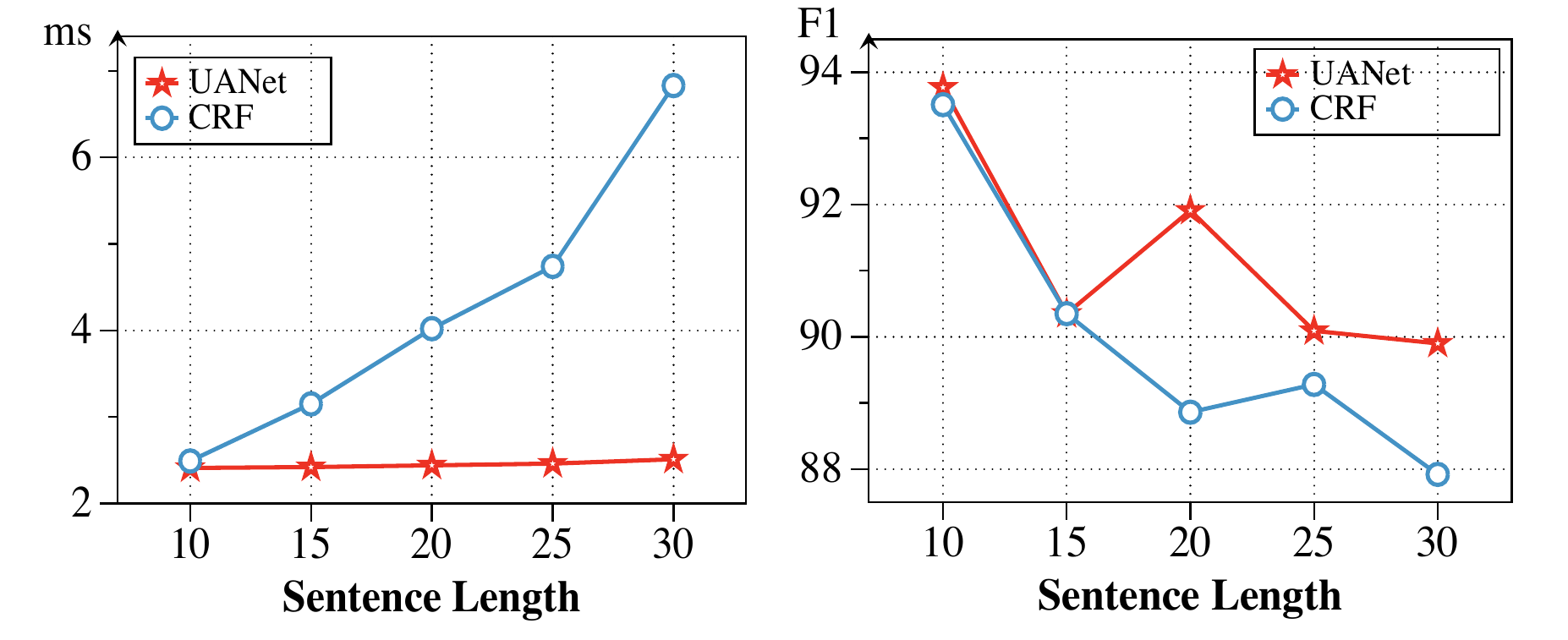}
  \caption{Speed and F1 against sentence length.} \label{fig:senlen}
\end{figure}

\begin{table*}[t]
\scriptsize
\begin{center}
\setlength{\tabcolsep}{1pt}
\begin{tabular}{|l|ccccccccccccccc|ccccc|}
\hline
\textbf{Text}             & ... & striker & Viorel & Ion   & of    & Otelul                       & Galati                       & and   & defender & Liviu & Ciobotariu & of    & National                     & Bucharest                    & ... & ... & University                   & of                           & Yangon                       & ... \\ \hline
\textbf{BiLSTM-CRF}       & ... & O       & B-PER  & E-PER & O     & \textbf{\color[HTML]{FE0000} B-PER} & \textbf{\color[HTML]{FE0000} E-PER} & O     & O        & B-PER & E-PER      & O     & \textbf{\color[HTML]{FE0000} B-LOC} & \textbf{\color[HTML]{FE0000} E-LOC} & ... & ... & \textbf{\color[HTML]{FE0000} O}     & \textbf{\color[HTML]{FE0000} O}     & \textbf{\color[HTML]{FE0000} S-LOC} & ... \\ \hline
\textbf{Draft Label}      & ... & O       & B-PER  & E-PER & O     & \textbf{\color[HTML]{FE0000} B-PER} & \textbf{\color[HTML]{FE0000} E-PER} & O     & O        & B-PER & E-PER      & O     & \textit{\color[HTML]{3166FF} B-ORG} & \textit{\color[HTML]{3166FF} E-ORG} & ... & ... & \textit{\color[HTML]{3166FF} B-ORG} & \textit{\color[HTML]{3166FF} I-ORG} & \textbf{\color[HTML]{FE0000} E-LOC} & ... \\
\textbf{Refinement}       & ... & O       & B-PER  & E-PER & O     & \textit{\color[HTML]{3166FF} B-ORG} & \textit{\color[HTML]{3166FF} E-ORG} & O     & O        & B-PER & E-PER      & O     & \textit{\color[HTML]{3166FF} B-ORG} & \textit{\color[HTML]{3166FF} E-ORG} & ... & ... & \textbf{\color[HTML]{FE0000} B-LOC} & \textit{\color[HTML]{3166FF} I-ORG} & \textit{\color[HTML]{3166FF} E-ORG} & ... \\
\textbf{Uncertainty}      & ... & 0.001   & 0.005  & 0.047 & 0.004 & \textit{\color[HTML]{3166FF} 0.532} & \textit{\color[HTML]{3166FF} 0.605} & 0.000 & 0.000    & 0.001 & 0.014      & 0.001 & \textit{\color[HTML]{3166FF} 0.818} & \textit{\color[HTML]{3166FF} 0.927} & ... & ... & {\color[HTML]{000000} 0.302} & \textit{\color[HTML]{3166FF} 0.816} & \textit{\color[HTML]{3166FF} 0.800} & ... \\ \hline
\textbf{Final Prediction} & ... & O       & B-PER  & E-PER & O     & B-ORG                        & E-ORG                        & O     & O        & B-PER & E-PER      & O     & B-ORG                        & E-ORG                        & ... & ... & B-ORG                        & I-ORG                        & E-ORG                        & ... \\ \hline

\end{tabular}
\end{center}
\caption{\label{tab:case} NER cases analysis. Contents with \textbf{bold red} and \textit{italic blue} styles represent incorrect and correct entities, respectively. Draft labels with uncertainty greater than 0.35 will be refined.}
\end{table*}

\subsection{Efficient Advantage}
Table \ref{tab:speed} shows a comparison of inference speeds. BiLSTM-UANet processes 1,630, 1,262, and 1,192 sentences per second on the CoNLL2003, OntoNotes, and WSJ development data, respectively, outperforming BiLSTM-CRF by 13.7\%, 32.8\% and 48.8\%, respectively. We can see that for the dataset with a longer average length, the speed of inference will be more advantageous. Because the model calculates uncertainties through parallel sampling the same input multiple times, the inference time of the BiLSTM-UANet ($M=8$) only slightly increases.

To further investigate the influence of the different sentence lengths, we analyze the inference speed of the UANet and CRF on the CoNLL2003 development set, which is split into five parts according to sentence length. We ruled out the influence of the text encoder and only counted the time of label decoding. The left subfigure in Figure \ref{fig:senlen} shows the decoding speed on the different sentence lengths. The results reveal that as the sentence length increases, the speed of the UANet is relatively stable, while the speed of the CRF decreases substantially. Due to the UANet's parallelism, when processing the sentence longer than 30, the UANet is nearly 3 times faster than the CRF. In addition, we exhibit the F1 score of the sentences with different lengths in right subfigure. It is worth noting that the UANet outperforms the CRF by a large margin when the length of the sentence is greater than 15, verifying the UANet's superiority in long-term label dependencies.

\begin{figure}[h]
\centering
  \includegraphics[width=3.1in]{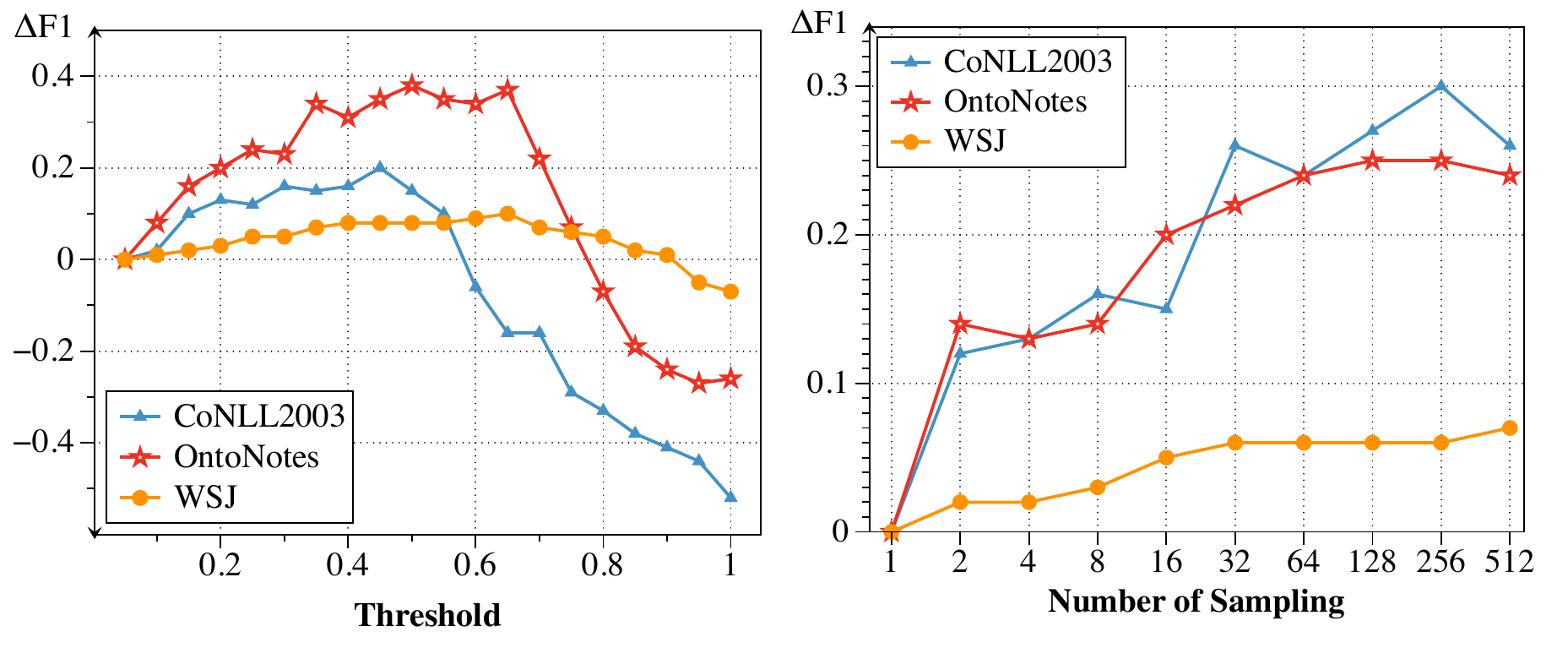}
  \caption{F1 variation under different uncertainty thresholds and numbers of sampling in variational LSTM, respectively. The results are evaluated on the development sets. $\Delta$F1 represents the F1 scores at different steps minus the initial results.} \label{fig:threshold}
\end{figure}

\subsection{Discussion}

\textbf{Uncertainty Threshold.} In order to investigate the influence of uncertainty threshold $\Gamma$, we evaluate the performance with different uncertainty thresholds on three datasets, as shown in Figure \ref{fig:threshold}. $\Gamma=0$ represents that the model uses all of the refined labels as final predictions. As the threshold gets larger, the performance of UANet can improve by reducing the negative effects on correct draft labels. However, when $\Gamma$ is too large, the model mainly uses draft labels as final predictions, resulting in performance degradation, which verifies our motivation that a reasonable uncertainty threshold can avoid side effects on correct draft labels. 


\textbf{Number of Sampling.} We also investigate the influence of the number of sampling in the variational LSTM as shown in Figure \ref{fig:threshold}. The results meet our expectation that a larger number of sampling can lead to better performance because a larger number of sampling can make the model better approximate the posterior $p(\mathbf{W}|\mathcal{D})$.

\textbf{Case Study.}
Table \ref{tab:case} shows two cases from CoNLL2003 NER dataset. The first case reflects the necessity of modeling higher-order dependencies in the NER task. UANet can learn the label consistency of two phrases near the word \texttt{and}. Moreover, seq2seq decoding model \cite{zhang2018learning} refines the labels in a left-to-right way and can't refine the previous labels in this case. The second case shows the effectiveness of the uncertainty threshold in mitigating the side effect of incorrect refinement. In this case, the refinement model is affected by the incorrect label of \texttt{Yangon} (E-LOC) when predicting the word \texttt{University}. Since the uncertainty value of \texttt{University} is lower than the threshold, our model can get the correct results. 


\section{Conclusions}
In this work, we introduce a novel sequence labeling framework that incorporates Bayesian neural networks to estimate model uncertainty. We find that the model uncertainty can effectively indicate the labels with a high probability of being wrong. The proposed method can selectively refine the uncertain labels to avoid the side effects of the refinement on correct labels. In addition, the proposed model can capture different ranges of label dependencies and word-label interactions in parallel, which can avoid the use of Viterbi decoding of the CRF for a faster prediction. Experimental results across three sequence labeling datasets demonstrated that the proposed method significantly outperforms the previous methods.

\section*{Acknowledgments}
The authors wish to thank the anonymous reviewers for their helpful comments. This work was partially funded by China National Key R\&D Program (No. 2018YFC0831105, 2018YFB1005104, 2017YFB1002104), National Natural Science Foundation of China (No. 61751201, 61976056, 61532011), Shanghai Municipal Science and Technology Major Project (No.2018SHZDZX01), Science and Technology Commission of Shanghai Municipality Grant  (No.18DZ1201000, 17JC1420200), Baidu Scholarship.

\bibliographystyle{acl_natbib}
\bibliography{anthology}

\end{document}